\pdfoutput=1

\documentclass[11pt]{article}

\usepackage[final]{acl}

\usepackage{times}
\usepackage{latexsym}

\usepackage[T1]{fontenc}

\usepackage[utf8]{inputenc}

\usepackage{microtype}

\usepackage{inconsolata}

\usepackage{graphicx}

\usepackage{enumitem}
\usepackage{geometry}

\usepackage{xcolor}
\usepackage{colortbl}
\usepackage{ulem} 
\usepackage{pifont}
\usepackage{booktabs}
\usepackage{amsmath}

\usepackage{soul}
\newcommand{\hlc}[2][yellow]{{%
    \colorlet{foo}{#1}%
    \sethlcolor{foo}\hl{#2}}%
}

\newcommand{\opentrack}[1]{\rowcolor{gray!20} #1}
\newcommand{\closedtrack}[1]{\rowcolor{gray!50} #1}
\newcommand{\nonsupporting}[1]{#1 \S}
\newcommand{\validated}{\colorbox{black}{\textcolor{white}{\ding{51}}}}

\usepackage[textsize=footnotesize]{todonotes}

\title{Preliminary WMT24 Ranking of General MT Systems and LLMs}

\author{
  \null \AND
  Tom Kocmi  
  \And
  Eleftherios Avramidis 
  \And
  Rachel Bawden 
  \And
  Ond\v{r}ej Bojar
  \AND
  Anton Dvorkovich 
  \And
  Christian Federmann  
  \And
  Mark Fishel  
  \And
  Markus Freitag
  \AND
  Thamme Gowda
  \And
  Roman Grundkiewicz
  \And
  Barry Haddow  
  \And
  Marzena Karpinska 
  \AND
  Philipp Koehn 
  \And
  Benjamin Marie 
  \And
  Kenton Murray 
  \And
  Masaaki Nagata
  \And
  Martin Popel 
  \AND 
  Maja Popovi\'{c}  
  \And
  Mariya Shmatova 
  \And
  Steinþór Steingrímsson 
  \And 
  Vilém Zouhar
  \vspace{2cm}
}

\begin{document}
\maketitle

\section*{Introduction}

This is the preliminary ranking of WMT24 General MT systems based on automatic metrics. The official ranking will be a human evaluation, which is superior to the automatic ranking and supersedes it.
The purpose of this report is not to interpret any findings but only provide preliminary results to the participants of the General MT task that may be useful during the writing of the system submission.
The automatic results are in the 11 pages following the main text. For the system submissions, use any of the tables without modifying them and subsampling systems or metrics.

\section*{Automatic Ranking}

Despite human ranking being superior to automatic ranking (see Limitations), we design a way of ranking systems before finishing the collection of human annotations.
Amidst the many options, we select two following metrics:
\begin{itemize}[noitemsep,left=0mm]
    \item MetricX-23-XL \citep{juraska-etal-2023-metricx} -- a reference based metric build on top of mT5 model.
    \item CometKiwi-DA-XL \citep{rei-etal-2023-scaling} -- a quality estimation metric built on the XLM-R XL model.
\end{itemize}

\noindent
Both metrics are top performing metrics \citep{freitag-etal-2023-results}.
We intentionally select two distinct metrics (underlying pretrained system and architecture) to minimize their bias and potential problems.
Although quality estimation is on average slightly worse than reference-based metrics, it helps us avoid a potential reference bias when human references are suboptimal \citep{freitag-etal-2023-results}.
Similarly, multilingual quality estimation can be fooled when the translation is in the incorrect language, which the reference-based metric would penalize.

To design the automatic ranking, we first linearly scale the scores of each metric to a range between 1 and \textit{the number of systems in a given language pair}.
After this, we average both normalized scores to arrive at the final automatic ranking, which we call AutoRank.

\section*{Types of Systems}
We distinguish three types of MT systems participating in the shared task:
\begin{itemize}
\item \textbf{Constrained systems} are those using only the specifically allowed training data for WMT2024 and the following pretrained models:
Llama-2-7B, Llama-2-13B, Mistral-7B, mBART, BERT, RoBERTa, XLM-RoBERTa, sBERT, LaBSE. Constrained systems may use any publicly available metric that was evaluated on past WMT Metrics shared tasks (e.g. COMET or Bleurt) and any basic linguistics tools (e.g. taggers, parsers, morphology analyzers).

\item \textbf{Open systems} (marked in tables with a \hlc[gray!20]{light gray} background) may use software and data under any open source license that places no constrains for non-commercial purposes (e.g. Apache, MIT) allowing to make the work replicable by any research group.

\item \textbf{Closed systems} (marked with \hlc[gray!50]{dark gray}) are all the remaining (fully automatic) systems, with no limitations on their training data (all ONLINE systems fit into this category).
\end{itemize}

\section*{Human Evaluation}

This year, we received an unusually high number of submitted systems and we are not able to evaluate them all with human annotators.
Therefore, we select a subset of 10 to 15 systems per language pair which will be evaluated by humans with the Error Span Annotation protocol \citep{kocmi2024errorspanannotationbalanced}.
For the remaining systems, AutoRank is going to be the official final ranking.

When selecting the systems for human evaluation, we prioritize open and constrained systems and penalize low-performing closed systems.
Therefore, we select the systems for human evaluation based on the following two rules:

First,
we exclude closed systems that are not among the first third of all systems
and we exclude open systems that are not among the top two thirds of all systems.

Second, motivated by several very low-performing systems, we also define a hard cutoff point.
After this point we do not evaluate any system from any category.
The cutoff point is selected as the first gap between two neighboring system's ranks larger than 1.5 of AutoRank.
We specify this cutoff point by a thick line in all the following tables in the rest of this paper.
The systems marked with $\dagger$ have been removed from human evaluation due to the author's request or not fulfilling submission requirements.

\section*{Evaluated Systems}

Details of all systems are going to be available in the upcoming WMT24 findings.
In addition to participants, we also collect most popular LLMs in a 3-shot approach, specifically, we collected: Aya23, Claude-3.5-Sonnet, Command R+, GPT-4, Gemini-1.5-Pro, Llama3-70B, Mistral-Large, and Phi-3-Medium.
We looked into the original reports for these LLMs to see which languages are claimed to be supported by checking if both source and target languages are mentioned or evaluated in any of their multilingual settings. 
We marked LLMs that do not officially claim a support for a given language by \nonsupporting{} in the tables. But to avoid the selection bias, we collect translations for all languages, even those not officially claimed to be supported.\footnote{The code for collecting translations: \\ \null\hfill \href{https://github.com/wmt-conference/wmt-collect-translations}{github.com/wmt-conference/wmt-collect-translations}}

\section*{Evaluated Data}

We evaluated 11 language pairs with approximately 1k segments per language (an exception is Czech$\rightarrow$Ukrainian with 2.3k segments). The average size is around 32k words per language pair.
The domains cover news, social, speech, and literary.
We do not provide any sentence splitting, thus many segments contain multiple sentences. 
We release all data at: \href{https://github.com/wmt-conference/wmt24-news-systems}{github.com/wmt-conference/wmt24-news-systems}.

\section*{Limitations}

Some models may have used Comet or MetricX during their training, for example, using Minimum Bayes Risk \citep{freitag-etal-2022-high}. Such models will be biased in this evaluation and will obtain artificially higher scores.

Automatic metrics are limited and biased \citep{karpinska-etal-2022-demetr, moghe2024machine}, which motivates the superseding by human evaluation. Another potential problem may have been that test sets we use are paragraph-level, however, automatic metrics have usually been tested in a sentence-level scenario.

We are not using BLEU nor ChrF on purpose, string-based metrics have for years low-performance when compared to humans \citep{kocmi-etal-2021-ship, freitag-etal-2022-results, freitag-etal-2023-results} and are especially misleading when comparing systems of different types \citet{callison-burch-etal-2006-evaluating, kocmi2024navigating}. Lastly, none of the evaluated languages is a low-resource language where pretrained metrics could struggle.

\section*{Acknowledgement}
This report would not have been possible without the partnership with Microsoft, Charles University, Dubformer, Toloka, NTT, Google, Árni Magnússon Institute, Custom.mt, Cohere, Together.ai, and Unbabel.
We are grateful for help from Nikolay Bogoychev and Konstantin Dranch.

\newcolumntype{L}[1]{>{\raggedright\let\newline\\\arraybackslash\hspace{0pt}}m{#1}}
\newcolumntype{C}[1]{>{\centering\let\newline\\\arraybackslash\hspace{0pt}}m{#1}}
\newcolumntype{R}[1]{>{\raggedleft\let\newline\\\arraybackslash\hspace{0pt}}m{#1}}

\clearpage
\begin{table*}
\centering
\begin{tabular}{c}
\bf{\Large{Czech-Ukrainian}}
\vspace{1em}
\end{tabular}
\begin{tabular}{R{40mm}C{22mm}C{19mm}C{22mm}C{32mm}}
\bf System Name & \bf AutoRank $\downarrow$ & \bf MetricX $\downarrow$ & \bf CometKiwi $\uparrow$ & \bf Human evaluation? \\
\toprule
\opentrack{IOL-Research & 1.9 & 1.3 & 0.681 & \validated} \\
\opentrack{IKUN & 2.3 & 1.6 & 0.664 & \validated} \\
\opentrack{Aya23 & 2.5 & 1.9 & 0.665 & \validated} \\
\opentrack{\nonsupporting{Llama3-70B} & 2.6 & 2.0 & 0.661 & } \\
CUNI-Transformer & 3.0 & 2.0 & 0.639 & \validated \\
IKUN-C & 3.0 & 2.4 & 0.648 & \validated \\
\midrule
CycleL & 21.0 & 19.5 & 0.146 &  \\
\bottomrule
\end{tabular}
\caption{Preliminary WMT24 General MT automatic ranking for Czech-Ukrainian (excluding closed systems).}
\vspace{2em}
\begin{tabular}{R{40mm}C{22mm}C{19mm}C{22mm}C{32mm}}
\bf System Name & \bf AutoRank $\downarrow$ & \bf MetricX $\downarrow$ & \bf CometKiwi $\uparrow$ & \bf Human evaluation? \\
\toprule
\closedtrack{Unbabel-Tower70B & 1.0 & 0.9 & 0.719 & \validated} \\
\closedtrack{\nonsupporting{Claude-3.5} & 1.7 & 1.0 & 0.683 & \validated} \\
\opentrack{IOL-Research & 1.9 & 1.3 & 0.681 & \validated} \\
\closedtrack{\nonsupporting{CommandR-plus} & 1.9 & 1.3 & 0.677 & \validated} \\
\closedtrack{\nonsupporting{GPT-4} & 2.0 & 1.4 & 0.677 & \validated} \\
\closedtrack{Gemini-1.5-Pro & 2.0 & 1.2 & 0.668 & \validated} \\
\closedtrack{ONLINE-W & 2.3 & 1.4 & 0.661 & \validated} \\
\closedtrack{\nonsupporting{Mistral-Large} & 2.3 & 1.6 & 0.666 & } \\
\opentrack{IKUN & 2.3 & 1.6 & 0.664 & \validated} \\
\opentrack{Aya23 & 2.5 & 1.9 & 0.665 & \validated} \\
\closedtrack{TranssionMT & 2.6 & 1.5 & 0.648 & } \\
\closedtrack{ONLINE-B & 2.6 & 1.6 & 0.648 & } \\
\closedtrack{ONLINE-A & 2.6 & 1.5 & 0.647 & } \\
\opentrack{\nonsupporting{Llama3-70B} & 2.6 & 2.0 & 0.661 & } \\
\closedtrack{ONLINE-G & 2.8 & 1.8 & 0.639 & } \\
CUNI-Transformer & 3.0 & 2.0 & 0.639 & \validated \\
IKUN-C & 3.0 & 2.4 & 0.648 & \validated \\
\midrule
\closedtrack{\nonsupporting{Phi-3-Medium} & 9.1 & 6.5 & 0.425 & } \\
\closedtrack{BJFU-LPT $\dagger$ & 11.5 & 7.6 & 0.321 & } \\
CycleL & 21.0 & 19.5 & 0.146 &  \\
\bottomrule
\end{tabular}
\caption{Preliminary WMT24 General MT automatic ranking for Czech-Ukrainian.}
\end{table*}

\clearpage
\begin{table*}
\centering
\begin{tabular}{c}
\bf{\Large{English-Czech}}
\vspace{1em}
\end{tabular}
\begin{tabular}{R{40mm}C{22mm}C{19mm}C{22mm}C{32mm}}
\bf System Name & \bf AutoRank $\downarrow$ & \bf MetricX $\downarrow$ & \bf CometKiwi $\uparrow$ & \bf Human evaluation? \\
\toprule
CUNI-MH & 2.1 & 2.3 & 0.690 & \validated \\
CUNI-GA & 2.3 & 3.7 & 0.726 & \validated \\
\opentrack{IOL-Research & 2.8 & 3.0 & 0.676 & \validated} \\
SCIR-MT & 3.2 & 3.3 & 0.664 & \validated \\
\opentrack{IKUN & 3.9 & 3.7 & 0.638 & \validated} \\
\opentrack{\nonsupporting{Llama3-70B} & 4.1 & 4.0 & 0.640 & \validated} \\
\opentrack{Aya23 & 4.3 & 4.0 & 0.630 & \validated} \\
CUNI-DocTransformer & 4.4 & 4.0 & 0.621 & \validated \\
IKUN-C & 4.7 & 4.3 & 0.618 & \validated \\
CUNI-Transformer $\dagger$ & 4.7 & 4.3 & 0.614 &  \\
\midrule
TSU-HITs & 19.5 & 16.6 & 0.235 &  \\
CycleL2 & 24.2 & 19.5 & 0.077 &  \\
CycleL & 27.0 & 22.5 & 0.031 &  \\
\bottomrule
\end{tabular}
\caption{Preliminary WMT24 General MT automatic ranking for English-Czech (excluding closed systems).}
\vspace{2em}
\begin{tabular}{R{40mm}C{22mm}C{19mm}C{22mm}C{32mm}}
\bf System Name & \bf AutoRank $\downarrow$ & \bf MetricX $\downarrow$ & \bf CometKiwi $\uparrow$ & \bf Human evaluation? \\
\toprule
\closedtrack{Unbabel-Tower70B & 1.0 & 1.8 & 0.732 & \validated} \\
\closedtrack{\nonsupporting{Claude-3.5} & 2.1 & 2.4 & 0.693 & \validated} \\
CUNI-MH & 2.1 & 2.3 & 0.690 & \validated \\
CUNI-GA & 2.3 & 3.7 & 0.726 & \validated \\
\closedtrack{Gemini-1.5-Pro & 2.6 & 2.8 & 0.678 & \validated} \\
\closedtrack{\nonsupporting{GPT-4} & 2.6 & 2.9 & 0.682 & \validated} \\
\opentrack{IOL-Research & 2.8 & 3.0 & 0.676 & \validated} \\
\closedtrack{ONLINE-W & 2.8 & 2.8 & 0.669 & \validated} \\
\closedtrack{\nonsupporting{CommandR-plus} & 2.9 & 2.9 & 0.669 & \validated} \\
SCIR-MT & 3.2 & 3.3 & 0.664 & \validated \\
\closedtrack{TranssionMT & 3.5 & 3.5 & 0.655 & } \\
\closedtrack{ONLINE-A & 3.6 & 3.4 & 0.648 & } \\
\closedtrack{\nonsupporting{Mistral-Large} & 3.7 & 3.6 & 0.647 & } \\
\opentrack{IKUN & 3.9 & 3.7 & 0.638 & \validated} \\
\closedtrack{ONLINE-B & 4.0 & 3.9 & 0.640 & } \\
\opentrack{\nonsupporting{Llama3-70B} & 4.1 & 4.0 & 0.640 & \validated} \\
\opentrack{Aya23 & 4.3 & 4.0 & 0.630 & \validated} \\
CUNI-DocTransformer & 4.4 & 4.0 & 0.621 & \validated \\
IKUN-C & 4.7 & 4.3 & 0.618 & \validated \\
CUNI-Transformer $\dagger$ & 4.7 & 4.3 & 0.614 &  \\
\closedtrack{ONLINE-G & 5.7 & 5.2 & 0.592 & } \\
\midrule
\closedtrack{NVIDIA-NeMo $\dagger$ & 7.6 & 6.5 & 0.536 & } \\
\closedtrack{\nonsupporting{Phi-3-Medium} & 15.0 & 11.4 & 0.305 & } \\
TSU-HITs & 19.5 & 16.6 & 0.235 &  \\
CycleL2 & 24.2 & 19.5 & 0.077 &  \\
CycleL & 27.0 & 22.5 & 0.031 &  \\
\bottomrule
\end{tabular}
\caption{Preliminary WMT24 General MT automatic ranking for English-Czech.}
\end{table*}

\clearpage
\begin{table*}
\centering
\begin{tabular}{c}
\bf{\Large{English-German}}
\vspace{1em}
\end{tabular}
\begin{tabular}{R{40mm}C{22mm}C{19mm}C{22mm}C{32mm}}
\bf System Name & \bf AutoRank $\downarrow$ & \bf MetricX $\downarrow$ & \bf CometKiwi $\uparrow$ & \bf Human evaluation? \\
\toprule
\opentrack{IOL-Research & 2.3 & 1.6 & 0.692 & \validated} \\
\opentrack{\nonsupporting{Llama3-70B} & 2.5 & 1.7 & 0.686 & \validated} \\
\opentrack{Aya23 & 2.7 & 1.8 & 0.680 & \validated} \\
\opentrack{IKUN & 3.0 & 1.8 & 0.668 & \validated} \\
IKUN-C & 3.8 & 2.0 & 0.641 & \validated \\
\opentrack{CUNI-NL & 4.2 & 2.1 & 0.624 & } \\
\midrule
AIST-AIRC & 7.2 & 3.3 & 0.551 &  \\
\opentrack{Occiglot & 8.2 & 3.8 & 0.539 & } \\
MSLC & 11.9 & 4.4 & 0.390 &  \\
TSU-HITs & 13.3 & 5.6 & 0.395 &  \\
CycleL2 & 27.0 & 11.5 & 0.091 &  \\
CycleL & 27.0 & 11.5 & 0.091 &  \\
\bottomrule
\end{tabular}
\caption{Preliminary WMT24 General MT automatic ranking for English-German (excluding closed systems).}
\vspace{2em}
\begin{tabular}{R{40mm}C{22mm}C{19mm}C{22mm}C{32mm}}
\bf System Name & \bf AutoRank $\downarrow$ & \bf MetricX $\downarrow$ & \bf CometKiwi $\uparrow$ & \bf Human evaluation? \\
\toprule
\closedtrack{Unbabel-Tower70B & 1.0 & 1.1 & 0.723 & \validated} \\
\closedtrack{Dubformer & 1.8 & 1.2 & 0.694 & \validated} \\
\closedtrack{TranssionMT & 1.8 & 1.4 & 0.699 & \validated} \\
\closedtrack{GPT-4 & 1.8 & 1.4 & 0.700 & \validated} \\
\closedtrack{ONLINE-B & 1.8 & 1.4 & 0.698 & \validated} \\
\closedtrack{Claude-3.5 & 1.9 & 1.4 & 0.695 & \validated} \\
\closedtrack{CommandR-plus & 2.0 & 1.4 & 0.696 & \validated} \\
\closedtrack{Mistral-Large & 2.0 & 1.5 & 0.694 & \validated} \\
\closedtrack{Gemini-1.5-Pro & 2.2 & 1.5 & 0.688 & \validated} \\
\closedtrack{ONLINE-W & 2.2 & 1.5 & 0.689 & } \\
\opentrack{IOL-Research & 2.3 & 1.6 & 0.692 & \validated} \\
\opentrack{\nonsupporting{Llama3-70B} & 2.5 & 1.7 & 0.686 & \validated} \\
\opentrack{Aya23 & 2.7 & 1.8 & 0.680 & \validated} \\
\opentrack{IKUN & 3.0 & 1.8 & 0.668 & \validated} \\
\closedtrack{ONLINE-A & 3.0 & 1.8 & 0.667 & } \\
\closedtrack{\nonsupporting{Phi-3-Medium} & 3.4 & 2.0 & 0.657 & } \\
\closedtrack{ONLINE-G & 3.5 & 2.1 & 0.662 & } \\
IKUN-C & 3.8 & 2.0 & 0.641 & \validated \\
\opentrack{CUNI-NL & 4.2 & 2.1 & 0.624 & } \\
\midrule
AIST-AIRC & 7.2 & 3.3 & 0.551 &  \\
\closedtrack{NVIDIA-NeMo $\dagger$ & 7.4 & 3.5 & 0.558 & } \\
\opentrack{Occiglot & 8.2 & 3.8 & 0.539 & } \\
MSLC & 11.9 & 4.4 & 0.390 &  \\
TSU-HITs & 13.3 & 5.6 & 0.395 &  \\
CycleL2 & 27.0 & 11.5 & 0.091 &  \\
CycleL & 27.0 & 11.5 & 0.091 &  \\
\bottomrule
\end{tabular}
\caption{Preliminary WMT24 General MT automatic ranking for English-German.}
\end{table*}

\clearpage
\begin{table*}
\centering
\begin{tabular}{c}
\bf{\Large{English-Spanish}}
\vspace{1em}
\end{tabular}
\begin{tabular}{R{40mm}C{22mm}C{19mm}C{22mm}C{32mm}}
\bf System Name & \bf AutoRank $\downarrow$ & \bf MetricX $\downarrow$ & \bf CometKiwi $\uparrow$ & \bf Human evaluation? \\
\toprule
\opentrack{IOL-Research & 2.3 & 2.8 & 0.701 & \validated} \\
\opentrack{\nonsupporting{Llama3-70B} & 2.6 & 3.0 & 0.693 & \validated} \\
\opentrack{IKUN & 2.8 & 3.3 & 0.687 & \validated} \\
\opentrack{Aya23 & 3.1 & 3.5 & 0.681 & } \\
IKUN-C & 3.4 & 3.5 & 0.666 & \validated \\
\opentrack{Occiglot & 5.9 & 5.4 & 0.583 & } \\
MSLC & 7.4 & 6.4 & 0.532 & \validated \\
\midrule
TSU-HITs & 16.3 & 14.2 & 0.289 &  \\
CycleL & 24.0 & 20.9 & 0.072 &  \\
\bottomrule
\end{tabular}
\caption{Preliminary WMT24 General MT automatic ranking for English-Spanish (excluding closed systems).}
\vspace{2em}
\begin{tabular}{R{40mm}C{22mm}C{19mm}C{22mm}C{32mm}}
\bf System Name & \bf AutoRank $\downarrow$ & \bf MetricX $\downarrow$ & \bf CometKiwi $\uparrow$ & \bf Human evaluation? \\
\toprule
\closedtrack{Unbabel-Tower70B & 1.0 & 1.9 & 0.745 & \validated} \\
\closedtrack{GPT-4 & 1.9 & 2.5 & 0.712 & \validated} \\
\closedtrack{Dubformer & 2.0 & 2.2 & 0.700 & \validated} \\
\closedtrack{CommandR-plus & 2.1 & 2.6 & 0.706 & \validated} \\
\closedtrack{Claude-3.5 & 2.1 & 2.6 & 0.705 & \validated} \\
\closedtrack{Mistral-Large & 2.2 & 2.7 & 0.707 & \validated} \\
\opentrack{IOL-Research & 2.3 & 2.8 & 0.701 & \validated} \\
\closedtrack{Gemini-1.5-Pro & 2.4 & 2.8 & 0.696 & \validated} \\
\opentrack{\nonsupporting{Llama3-70B} & 2.6 & 3.0 & 0.693 & \validated} \\
\closedtrack{ONLINE-B & 2.7 & 3.1 & 0.690 & } \\
\closedtrack{ONLINE-W & 2.7 & 3.0 & 0.682 & } \\
\closedtrack{TranssionMT & 2.8 & 3.2 & 0.689 & } \\
\opentrack{IKUN & 2.8 & 3.3 & 0.687 & \validated} \\
\closedtrack{\nonsupporting{Phi-3-Medium} & 3.0 & 3.4 & 0.685 & } \\
\closedtrack{ONLINE-A & 3.0 & 3.3 & 0.676 & } \\
\opentrack{Aya23 & 3.1 & 3.5 & 0.681 & } \\
\closedtrack{ONLINE-G & 3.2 & 3.6 & 0.674 & } \\
IKUN-C & 3.4 & 3.5 & 0.666 & \validated \\
\closedtrack{NVIDIA-NeMo $\dagger$ & 4.5 & 4.4 & 0.631 & } \\
\opentrack{Occiglot & 5.9 & 5.4 & 0.583 & } \\
MSLC & 7.4 & 6.4 & 0.532 & \validated \\
\midrule
TSU-HITs & 16.3 & 14.2 & 0.289 &  \\
CycleL & 24.0 & 20.9 & 0.072 &  \\
\bottomrule
\end{tabular}
\caption{Preliminary WMT24 General MT automatic ranking for English-Spanish.}
\end{table*}

\clearpage
\begin{table*}
\centering
\begin{tabular}{c}
\bf{\Large{English-Hindi}}
\vspace{1em}
\end{tabular}
\begin{tabular}{R{40mm}C{22mm}C{19mm}C{22mm}C{32mm}}
\bf System Name & \bf AutoRank $\downarrow$ & \bf MetricX $\downarrow$ & \bf CometKiwi $\uparrow$ & \bf Human evaluation? \\
\toprule
\opentrack{IOL-Research & 2.1 & 4.3 & 0.622 & \validated} \\
\opentrack{\nonsupporting{Llama3-70B} & 2.1 & 4.6 & 0.630 & \validated} \\
\opentrack{Aya23 & 3.2 & 5.4 & 0.591 & \validated} \\
IKUN-C & 5.5 & 7.1 & 0.499 & \validated \\
\midrule
\opentrack{IKUN & 7.7 & 9.4 & 0.428 & } \\
CycleL & 20.0 & 23.4 & 0.083 &  \\
\bottomrule
\end{tabular}
\caption{Preliminary WMT24 General MT automatic ranking for English-Hindi (excluding closed systems).}
\vspace{2em}
\begin{tabular}{R{40mm}C{22mm}C{19mm}C{22mm}C{32mm}}
\bf System Name & \bf AutoRank $\downarrow$ & \bf MetricX $\downarrow$ & \bf CometKiwi $\uparrow$ & \bf Human evaluation? \\
\toprule
\closedtrack{Unbabel-Tower70B & 1.0 & 3.1 & 0.657 & \validated} \\
\closedtrack{\nonsupporting{Claude-3.5} & 1.2 & 3.3 & 0.649 & \validated} \\
\closedtrack{TranssionMT & 1.3 & 3.3 & 0.644 & \validated} \\
\closedtrack{ONLINE-B & 1.4 & 3.3 & 0.641 & \validated} \\
\closedtrack{\nonsupporting{Gemini-1.5-Pro} & 1.6 & 3.6 & 0.635 & \validated} \\
\closedtrack{\nonsupporting{GPT-4} & 2.1 & 4.5 & 0.628 & \validated} \\
\opentrack{IOL-Research & 2.1 & 4.3 & 0.622 & \validated} \\
\opentrack{\nonsupporting{Llama3-70B} & 2.1 & 4.6 & 0.630 & \validated} \\
\closedtrack{\nonsupporting{CommandR-plus} & 2.3 & 4.4 & 0.612 & } \\
\opentrack{Aya23 & 3.2 & 5.4 & 0.591 & \validated} \\
\closedtrack{ONLINE-A & 3.5 & 6.2 & 0.590 & } \\
\closedtrack{ONLINE-G & 4.2 & 7.4 & 0.583 & } \\
\closedtrack{\nonsupporting{Mistral-Large} & 5.0 & 7.7 & 0.541 & } \\
IKUN-C & 5.5 & 7.1 & 0.499 & \validated \\
\closedtrack{NVIDIA-NeMo $\dagger$ & 5.8 & 8.9 & 0.530 & } \\
\midrule
\closedtrack{\nonsupporting{Phi-3-Medium} & 7.4 & 10.7 & 0.483 & } \\
\opentrack{IKUN & 7.7 & 9.4 & 0.428 & } \\
\closedtrack{ONLINE-W & 15.3 & 20.9 & 0.296 & } \\
CycleL & 20.0 & 23.4 & 0.083 &  \\
\bottomrule
\end{tabular}
\caption{Preliminary WMT24 General MT automatic ranking for English-Hindi.}
\end{table*}

\clearpage
\begin{table*}
\centering
\begin{tabular}{c}
\bf{\Large{English-Icelandic}}
\vspace{1em}
\end{tabular}
\begin{tabular}{R{40mm}C{22mm}C{19mm}C{22mm}C{32mm}}
\bf System Name & \bf AutoRank $\downarrow$ & \bf MetricX $\downarrow$ & \bf CometKiwi $\uparrow$ & \bf Human evaluation? \\
\toprule
\opentrack{IKUN & 3.2 & 4.3 & 0.666 & \validated} \\
\opentrack{AMI & 3.7 & 4.9 & 0.663 & \validated} \\
IKUN-C & 3.7 & 4.9 & 0.657 & \validated \\
\opentrack{IOL-Research & 4.3 & 5.7 & 0.655 & \validated} \\
\opentrack{\nonsupporting{Llama3-70B} & 6.7 & 8.0 & 0.586 & \validated} \\
\midrule
\opentrack{\nonsupporting{Aya23} & 15.2 & 14.9 & 0.311 & } \\
TSU-HITs & 19.2 & 18.4 & 0.192 &  \\
CycleL & 21.0 & 20.2 & 0.148 &  \\
\bottomrule
\end{tabular}
\caption{Preliminary WMT24 General MT automatic ranking for English-Icelandic (excluding closed systems).}
\vspace{2em}
\begin{tabular}{R{40mm}C{22mm}C{19mm}C{22mm}C{32mm}}
\bf System Name & \bf AutoRank $\downarrow$ & \bf MetricX $\downarrow$ & \bf CometKiwi $\uparrow$ & \bf Human evaluation? \\
\toprule
\closedtrack{Unbabel-Tower70B & 1.0 & 2.5 & 0.740 & \validated} \\
\closedtrack{\nonsupporting{Claude-3.5} & 2.3 & 3.6 & 0.697 & \validated} \\
\closedtrack{Dubformer & 2.5 & 3.4 & 0.685 & \validated} \\
\opentrack{IKUN & 3.2 & 4.3 & 0.666 & \validated} \\
\closedtrack{GPT-4 & 3.4 & 4.7 & 0.673 & \validated} \\
\opentrack{AMI & 3.7 & 4.9 & 0.663 & \validated} \\
IKUN-C & 3.7 & 4.9 & 0.657 & \validated \\
\closedtrack{TranssionMT & 4.2 & 5.5 & 0.653 & } \\
\closedtrack{ONLINE-B & 4.2 & 5.5 & 0.652 & } \\
\opentrack{IOL-Research & 4.3 & 5.7 & 0.655 & \validated} \\
\closedtrack{ONLINE-A & 5.5 & 6.4 & 0.603 & } \\
\opentrack{\nonsupporting{Llama3-70B} & 6.7 & 8.0 & 0.586 & \validated} \\
\closedtrack{ONLINE-G & 6.9 & 7.9 & 0.573 & } \\
\midrule
\closedtrack{\nonsupporting{CommandR-plus} & 9.8 & 10.6 & 0.487 & } \\
\closedtrack{\nonsupporting{Mistral-Large} & 10.4 & 10.9 & 0.465 & } \\
\opentrack{\nonsupporting{Aya23} & 15.2 & 14.9 & 0.311 & } \\
\closedtrack{\nonsupporting{Phi-3-Medium} & 16.2 & 15.7 & 0.278 & } \\
\closedtrack{ONLINE-W & 18.1 & 19.5 & 0.296 & } \\
TSU-HITs & 19.2 & 18.4 & 0.192 &  \\
CycleL & 21.0 & 20.2 & 0.148 &  \\
\bottomrule
\end{tabular}
\caption{Preliminary WMT24 General MT automatic ranking for English-Icelandic.}
\end{table*}

\clearpage
\begin{table*}
\centering
\begin{tabular}{c}
\bf{\Large{English-Japanese}}
\vspace{1em}
\end{tabular}
\begin{tabular}{R{40mm}C{22mm}C{19mm}C{22mm}C{32mm}}
\bf System Name & \bf AutoRank $\downarrow$ & \bf MetricX $\downarrow$ & \bf CometKiwi $\uparrow$ & \bf Human evaluation? \\
\toprule
Team-J & 1.9 & 2.9 & 0.740 & \validated \\
NTTSU & 1.9 & 2.6 & 0.731 & \validated \\
\opentrack{IOL-Research & 2.3 & 3.1 & 0.724 & \validated} \\
\opentrack{Aya23 & 2.3 & 3.1 & 0.719 & \validated} \\
\opentrack{\nonsupporting{Llama3-70B} & 2.6 & 3.5 & 0.714 & \validated} \\
\opentrack{IKUN & 3.1 & 3.7 & 0.696 & } \\
IKUN-C & 3.9 & 4.3 & 0.669 & \validated \\
\midrule
AIST-AIRC & 6.6 & 6.5 & 0.583 &  \\
CycleL & 24.0 & 22.4 & 0.101 &  \\
\bottomrule
\end{tabular}
\caption{Preliminary WMT24 General MT automatic ranking for English-Japanese (excluding closed systems).}
\vspace{2em}
\begin{tabular}{R{40mm}C{22mm}C{19mm}C{22mm}C{32mm}}
\bf System Name & \bf AutoRank $\downarrow$ & \bf MetricX $\downarrow$ & \bf CometKiwi $\uparrow$ & \bf Human evaluation? \\
\toprule
\closedtrack{Unbabel-Tower70B & 1.0 & 2.0 & 0.762 & \validated} \\
\closedtrack{ONLINE-B & 1.4 & 2.4 & 0.750 & \validated} \\
\closedtrack{Claude-3.5 & 1.5 & 2.3 & 0.744 & \validated} \\
\closedtrack{Gemini-1.5-Pro & 1.7 & 2.5 & 0.734 & \validated} \\
\closedtrack{GPT-4 & 1.7 & 2.7 & 0.740 & \validated} \\
Team-J & 1.9 & 2.9 & 0.740 & \validated \\
NTTSU & 1.9 & 2.6 & 0.731 & \validated \\
\closedtrack{CommandR-plus & 1.9 & 2.7 & 0.730 & \validated} \\
\opentrack{IOL-Research & 2.3 & 3.1 & 0.724 & \validated} \\
\opentrack{Aya23 & 2.3 & 3.1 & 0.719 & \validated} \\
\opentrack{\nonsupporting{Llama3-70B} & 2.6 & 3.5 & 0.714 & \validated} \\
\closedtrack{DLUT-GTCOM & 2.6 & 3.0 & 0.697 & } \\
\closedtrack{\nonsupporting{Phi-3-Medium} & 2.8 & 3.6 & 0.709 & } \\
\closedtrack{ONLINE-W & 2.9 & 3.6 & 0.700 & } \\
\closedtrack{\nonsupporting{Mistral-Large} & 2.9 & 3.8 & 0.707 & } \\
\closedtrack{ONLINE-A & 3.0 & 3.6 & 0.699 & } \\
\opentrack{IKUN & 3.1 & 3.7 & 0.696 & } \\
IKUN-C & 3.9 & 4.3 & 0.669 & \validated \\
\midrule
\closedtrack{ONLINE-G & 6.4 & 6.6 & 0.599 & } \\
AIST-AIRC & 6.6 & 6.5 & 0.583 &  \\
\closedtrack{UvA-MT & 6.7 & 6.7 & 0.589 & } \\
\closedtrack{NVIDIA-NeMo $\dagger$ & 6.9 & 6.9 & 0.582 & } \\
CycleL & 24.0 & 22.4 & 0.101 &  \\
\bottomrule
\end{tabular}
\caption{Preliminary WMT24 General MT automatic ranking for English-Japanese.}
\end{table*}

\clearpage
\begin{table*}
\centering
\begin{tabular}{c}
\bf{\Large{English-Russian}}
\vspace{1em}
\end{tabular}
\begin{tabular}{R{40mm}C{22mm}C{19mm}C{22mm}C{32mm}}
\bf System Name & \bf AutoRank $\downarrow$ & \bf MetricX $\downarrow$ & \bf CometKiwi $\uparrow$ & \bf Human evaluation? \\
\toprule
\opentrack{IOL-Research & 2.6 & 3.7 & 0.694 & \validated} \\
\opentrack{\nonsupporting{Llama3-70B} & 3.1 & 4.1 & 0.681 & \validated} \\
\opentrack{IKUN & 3.2 & 4.1 & 0.675 & \validated} \\
\opentrack{Aya23 & 3.3 & 4.2 & 0.669 & \validated} \\
IKUN-C & 3.9 & 4.7 & 0.649 & \validated \\
\midrule
CUNI-DS & 5.9 & 6.2 & 0.584 &  \\
TSU-HITs & 10.8 & 9.8 & 0.421 &  \\
CycleL & 24.3 & 22.2 & 0.062 &  \\
CycleL2 & 25.0 & 22.4 & 0.027 &  \\
\bottomrule
\end{tabular}
\caption{Preliminary WMT24 General MT automatic ranking for English-Russian (excluding closed systems).}
\vspace{2em}
\begin{tabular}{R{40mm}C{22mm}C{19mm}C{22mm}C{32mm}}
\bf System Name & \bf AutoRank $\downarrow$ & \bf MetricX $\downarrow$ & \bf CometKiwi $\uparrow$ & \bf Human evaluation? \\
\toprule
\closedtrack{Unbabel-Tower70B & 1.0 & 2.4 & 0.742 & \validated} \\
\closedtrack{Dubformer & 1.9 & 2.8 & 0.701 & \validated} \\
\closedtrack{Yandex & 1.9 & 2.9 & 0.705 & \validated} \\
\closedtrack{Claude-3.5 & 2.0 & 3.0 & 0.706 & \validated} \\
\closedtrack{ONLINE-G & 2.2 & 3.3 & 0.706 & \validated} \\
\closedtrack{GPT-4 & 2.3 & 3.4 & 0.703 & \validated} \\
\closedtrack{Gemini-1.5-Pro & 2.3 & 3.2 & 0.697 & \validated} \\
\closedtrack{\nonsupporting{CommandR-plus} & 2.4 & 3.4 & 0.693 & \validated} \\
\closedtrack{ONLINE-W & 2.6 & 3.5 & 0.688 & } \\
\opentrack{IOL-Research & 2.6 & 3.7 & 0.694 & \validated} \\
\closedtrack{\nonsupporting{Mistral-Large} & 2.7 & 3.7 & 0.692 & } \\
\opentrack{\nonsupporting{Llama3-70B} & 3.1 & 4.1 & 0.681 & \validated} \\
\closedtrack{ONLINE-B & 3.1 & 3.9 & 0.673 & } \\
\closedtrack{TranssionMT & 3.1 & 3.9 & 0.673 & } \\
\opentrack{IKUN & 3.2 & 4.1 & 0.675 & \validated} \\
\opentrack{Aya23 & 3.3 & 4.2 & 0.669 & \validated} \\
\closedtrack{ONLINE-A & 3.4 & 4.1 & 0.663 & } \\
\closedtrack{\nonsupporting{Phi-3-Medium} & 3.9 & 4.7 & 0.654 & } \\
IKUN-C & 3.9 & 4.7 & 0.649 & \validated \\
\midrule
CUNI-DS & 5.9 & 6.2 & 0.584 &  \\
\closedtrack{NVIDIA-NeMo $\dagger$ & 7.2 & 7.3 & 0.549 & } \\
TSU-HITs & 10.8 & 9.8 & 0.421 &  \\
CycleL & 24.3 & 22.2 & 0.062 &  \\
CycleL2 & 25.0 & 22.4 & 0.027 &  \\
\bottomrule
\end{tabular}
\caption{Preliminary WMT24 General MT automatic ranking for English-Russian.}
\end{table*}

\clearpage
\begin{table*}
\centering
\begin{tabular}{c}
\bf{\Large{English-Ukrainian}}
\vspace{1em}
\end{tabular}
\begin{tabular}{R{40mm}C{22mm}C{19mm}C{22mm}C{32mm}}
\bf System Name & \bf AutoRank $\downarrow$ & \bf MetricX $\downarrow$ & \bf CometKiwi $\uparrow$ & \bf Human evaluation? \\
\toprule
\opentrack{IOL-Research & 2.4 & 3.4 & 0.675 & \validated} \\
\opentrack{IKUN & 2.8 & 3.7 & 0.661 & \validated} \\
\opentrack{\nonsupporting{Llama3-70B} & 3.2 & 4.2 & 0.647 & } \\
\opentrack{Aya23 & 3.3 & 4.2 & 0.642 & } \\
IKUN-C & 3.9 & 4.7 & 0.622 & \validated \\
\midrule
CycleL & 21.0 & 22.4 & 0.037 &  \\
\bottomrule
\end{tabular}
\caption{Preliminary WMT24 General MT automatic ranking for English-Ukrainian (excluding closed systems).}
\vspace{2em}
\begin{tabular}{R{40mm}C{22mm}C{19mm}C{22mm}C{32mm}}
\bf System Name & \bf AutoRank $\downarrow$ & \bf MetricX $\downarrow$ & \bf CometKiwi $\uparrow$ & \bf Human evaluation? \\
\toprule
\closedtrack{Unbabel-Tower70B & 1.0 & 2.2 & 0.732 & \validated} \\
\closedtrack{Dubformer & 1.8 & 2.7 & 0.691 & \validated} \\
\closedtrack{Claude-3.5 & 2.0 & 3.0 & 0.693 & \validated} \\
\closedtrack{ONLINE-W & 2.1 & 2.8 & 0.679 & \validated} \\
\closedtrack{Gemini-1.5-Pro & 2.2 & 3.0 & 0.677 & \validated} \\
\closedtrack{\nonsupporting{CommandR-plus} & 2.3 & 3.2 & 0.678 & \validated} \\
\closedtrack{GPT-4 & 2.3 & 3.3 & 0.682 & \validated} \\
\closedtrack{ONLINE-G & 2.3 & 3.1 & 0.670 & } \\
\opentrack{IOL-Research & 2.4 & 3.4 & 0.675 & \validated} \\
\closedtrack{\nonsupporting{Mistral-Large} & 2.4 & 3.4 & 0.675 & } \\
\opentrack{IKUN & 2.8 & 3.7 & 0.661 & \validated} \\
\closedtrack{ONLINE-B & 3.1 & 3.9 & 0.646 & } \\
\closedtrack{TranssionMT & 3.1 & 4.0 & 0.646 & } \\
\opentrack{\nonsupporting{Llama3-70B} & 3.2 & 4.2 & 0.647 & } \\
\opentrack{Aya23 & 3.3 & 4.2 & 0.642 & } \\
\closedtrack{ONLINE-A & 3.3 & 4.1 & 0.634 & } \\
IKUN-C & 3.9 & 4.7 & 0.622 & \validated \\
\midrule
\closedtrack{NVIDIA-NeMo $\dagger$ & 6.2 & 7.0 & 0.537 & } \\
\closedtrack{\nonsupporting{Phi-3-Medium} & 11.1 & 11.3 & 0.339 & } \\
CycleL & 21.0 & 22.4 & 0.037 &  \\
\bottomrule
\end{tabular}
\caption{Preliminary WMT24 General MT automatic ranking for English-Ukrainian.}
\end{table*}

\clearpage
\begin{table*}
\centering
\begin{tabular}{c}
\bf{\Large{English-Chinese}}
\vspace{1em}
\end{tabular}
\begin{tabular}{R{40mm}C{22mm}C{19mm}C{22mm}C{32mm}}
\bf System Name & \bf AutoRank $\downarrow$ & \bf MetricX $\downarrow$ & \bf CometKiwi $\uparrow$ & \bf Human evaluation? \\
\toprule
\opentrack{IOL-Research & 1.8 & 3.1 & 0.700 & \validated} \\
HW-TSC & 2.3 & 3.4 & 0.675 & \validated \\
\opentrack{\nonsupporting{Llama3-70B} & 2.8 & 3.9 & 0.662 & \validated} \\
\opentrack{Aya23 & 3.0 & 4.1 & 0.655 & \validated} \\
\opentrack{IKUN & 3.1 & 4.0 & 0.646 & \validated} \\
IKUN-C & 3.5 & 4.2 & 0.624 & \validated \\
\midrule
CycleL & 20.1 & 20.1 & 0.086 &  \\
CycleL2 & 22.0 & 22.1 & 0.030 &  \\
\bottomrule
\end{tabular}
\caption{Preliminary WMT24 General MT automatic ranking for English-Chinese (excluding closed systems).}
\vspace{2em}
\begin{tabular}{R{40mm}C{22mm}C{19mm}C{22mm}C{32mm}}
\bf System Name & \bf AutoRank $\downarrow$ & \bf MetricX $\downarrow$ & \bf CometKiwi $\uparrow$ & \bf Human evaluation? \\
\toprule
\closedtrack{Unbabel-Tower70B & 1.0 & 2.3 & 0.726 & \validated} \\
\closedtrack{Claude-3.5 & 1.7 & 3.0 & 0.703 & \validated} \\
\closedtrack{ONLINE-B & 1.7 & 2.9 & 0.697 & \validated} \\
\opentrack{IOL-Research & 1.8 & 3.1 & 0.700 & \validated} \\
\closedtrack{Gemini-1.5-Pro & 1.8 & 3.1 & 0.698 & \validated} \\
\closedtrack{GPT-4 & 2.0 & 3.3 & 0.693 & \validated} \\
\closedtrack{CommandR-plus & 2.2 & 3.3 & 0.681 & \validated} \\
\closedtrack{ONLINE-W & 2.2 & 3.2 & 0.677 & } \\
HW-TSC & 2.3 & 3.4 & 0.675 & \validated \\
\closedtrack{\nonsupporting{Mistral-Large} & 2.8 & 4.0 & 0.665 & } \\
\opentrack{\nonsupporting{Llama3-70B} & 2.8 & 3.9 & 0.662 & \validated} \\
\opentrack{Aya23 & 3.0 & 4.1 & 0.655 & \validated} \\
\opentrack{IKUN & 3.1 & 4.0 & 0.646 & \validated} \\
\closedtrack{\nonsupporting{Phi-3-Medium} & 3.1 & 4.2 & 0.648 & } \\
\closedtrack{ONLINE-A & 3.3 & 4.1 & 0.636 & } \\
IKUN-C & 3.5 & 4.2 & 0.624 & \validated \\
\closedtrack{UvA-MT & 4.3 & 5.2 & 0.607 & } \\
\closedtrack{ONLINE-G & 4.8 & 5.5 & 0.588 & } \\
\midrule
\closedtrack{NVIDIA-NeMo $\dagger$ & 7.3 & 7.6 & 0.494 & } \\
CycleL & 20.1 & 20.1 & 0.086 &  \\
CycleL2 & 22.0 & 22.1 & 0.030 &  \\
\bottomrule
\end{tabular}
\caption{Preliminary WMT24 General MT automatic ranking for English-Chinese.}
\end{table*}

\clearpage
\begin{table*}
\centering
\begin{tabular}{c}
\bf{\Large{Japanese-Chinese}}
\vspace{1em}
\end{tabular}
\begin{tabular}{R{40mm}C{22mm}C{19mm}C{22mm}C{32mm}}
\bf System Name & \bf AutoRank $\downarrow$ & \bf MetricX $\downarrow$ & \bf CometKiwi $\uparrow$ & \bf Human evaluation? \\
\toprule
\opentrack{IOL-Research & 2.2 & 3.9 & 0.593 & \validated} \\
Team-J & 2.8 & 4.0 & 0.570 & \validated \\
\opentrack{\nonsupporting{Llama3-70B} & 3.1 & 4.7 & 0.578 & \validated} \\
\opentrack{Aya23 & 3.7 & 5.0 & 0.563 & \validated} \\
\opentrack{NTTSU & 3.7 & 5.3 & 0.566 & \validated} \\
\opentrack{IKUN & 4.4 & 5.4 & 0.544 & \validated} \\
IKUN-C & 5.5 & 6.2 & 0.519 & \validated \\
\midrule
MSLC & 8.9 & 8.8 & 0.450 &  \\
CycleL & 23.0 & 21.5 & 0.202 &  \\
\bottomrule
\end{tabular}
\caption{Preliminary WMT24 General MT automatic ranking for Japanese-Chinese (excluding closed systems).}
\vspace{2em}
\begin{tabular}{R{40mm}C{22mm}C{19mm}C{22mm}C{32mm}}
\bf System Name & \bf AutoRank $\downarrow$ & \bf MetricX $\downarrow$ & \bf CometKiwi $\uparrow$ & \bf Human evaluation? \\
\toprule
\closedtrack{Unbabel-Tower70B & 1.0 & 3.2 & 0.622 & \validated} \\
\closedtrack{Claude-3.5 & 1.7 & 3.5 & 0.603 & \validated} \\
\closedtrack{Gemini-1.5-Pro & 1.9 & 3.5 & 0.595 & \validated} \\
\closedtrack{DLUT-GTCOM & 2.0 & 3.3 & 0.586 & \validated} \\
\closedtrack{GPT-4 & 2.1 & 3.8 & 0.597 & \validated} \\
\opentrack{IOL-Research & 2.2 & 3.9 & 0.593 & \validated} \\
\closedtrack{CommandR-plus & 2.8 & 4.1 & 0.576 & \validated} \\
Team-J & 2.8 & 4.0 & 0.570 & \validated \\
\opentrack{\nonsupporting{Llama3-70B} & 3.1 & 4.7 & 0.578 & \validated} \\
\closedtrack{\nonsupporting{Mistral-Large} & 3.5 & 4.9 & 0.568 & } \\
\opentrack{Aya23 & 3.7 & 5.0 & 0.563 & \validated} \\
\opentrack{NTTSU & 3.7 & 5.3 & 0.566 & \validated} \\
\closedtrack{\nonsupporting{Phi-3-Medium} & 4.0 & 5.1 & 0.552 & } \\
\opentrack{IKUN & 4.4 & 5.4 & 0.544 & \validated} \\
\closedtrack{ONLINE-B & 5.2 & 5.5 & 0.518 & } \\
\closedtrack{UvA-MT & 5.2 & 6.3 & 0.534 & } \\
\closedtrack{ONLINE-W & 5.3 & 6.0 & 0.522 & } \\
IKUN-C & 5.5 & 6.2 & 0.519 & \validated \\
\closedtrack{ONLINE-A & 6.8 & 6.8 & 0.484 & } \\
\midrule
MSLC & 8.9 & 8.8 & 0.450 &  \\
\closedtrack{ONLINE-G & 10.3 & 9.6 & 0.413 & } \\
CycleL & 23.0 & 21.5 & 0.202 &  \\
\bottomrule
\end{tabular}
\caption{Preliminary WMT24 General MT automatic ranking for Japanese-Chinese.}
\end{table*}

\clearpage
\bibliography{anthology.min.bib,custom}

\end{document}